\documentclass[journal]{IEEEtran}
\usepackage{cite}
\usepackage{amsmath,amssymb,amsfonts}
\usepackage{algorithmic}
\usepackage{graphicx}
\usepackage{algorithm,algorithmic}
\usepackage{hyperref}
\hypersetup{hidelinks=true}
\usepackage{textcomp}
\usepackage{verbatim}
\usepackage{amsmath}
\usepackage{booktabs}
\usepackage{multirow}
\usepackage{amssymb}   
\usepackage{graphicx} 
\usepackage{placeins}
\usepackage{caption}
\usepackage{float}

\def\BibTeX{{\rm B\kern-.05em{\sc i\kern-.025em b}\kern-.08em
    T\kern-.1667em\lower.7ex\hbox{E}\kern-.125emX}}
\begin{document}
\title{MDSkin-Net: Multi-Task Skin Lesion Analysis Driven by 
Pattern Analysis Priors and Spatial Alignment Regularization}
\author{
Yijian~Li,
Saad~Bedros,
Paul~Bigliardi,
Mei~Bigliardi~Qi,
Vassilios~Morellas,
and Nikolaos~Papanikolopoulos,~\IEEEmembership{Fellow,~IEEE}%
\thanks{Y. Li, S. Bedros, V. Morellas, and N. Papanikolopoulos are with the University of Minnesota, Minneapolis, MN, USA.}%
\thanks{Corresponding author: Nikolaos Papanikolopoulos (email: papan001@umn.edu).}
}

\maketitle
\begin{abstract}
Reliable skin lesion segmentation and classification are central to dermoscopic computer-aided diagnosis. Existing multi-task frameworks couple the two tasks architecturally without clinical knowledge, while knowledge-injecting approaches rely on the macroscopic ABCD rule, which was not designed for dermoscopy. Dermoscopic diagnosis is grounded in Pattern Analysis, a microscopic framework structured around dermoscopic features. We propose MDSkin-Net, which incorporates cue-level Pattern Analysis priors into a hybrid CNN-Transformer architecture. At its core is a Pattern Analysis-Guided Attention Module (PAGAM) comprising three priors motivated by distinct dermoscopic cues: an improved Efficient Channel Attention (iECA), a Multi-Scale Spatial Attention (MSSA), and a Biased Asymmetry Attention (BAA). We further introduce a multi-scale spatial alignment regularization (MSAR) that uses the segmentation ground-truth mask as hierarchical soft supervision, confining the classification head to lesion-localized evidence and coupling both task pathways through a shared spatial prior. Trained exclusively on the ISIC 2017 training split without external dermoscopy data, the MDSkin-Net ensemble transfers robustly under zero-shot evaluation, reaching a Dice Similarity Coefficient (DSC) of 92.38\% and a melanoma AUC of 97.84\% on PH2, and a DSC of 88.92\% on the ISIC 2018 Task 1 test set. On the in-domain ISIC 2017 benchmark, the ensemble attains a mean Area Under the Curve (AUC) of 91.60\% across the two classification tasks (melanoma and seborrheic keratosis vs. rest), and a DSC of 84.72\% for segmentation. Classification remains competitive with baselines; in-domain segmentation trails single-task specialists, yet the proposed priors and alignment regularization yield representations that generalize consistently across cohorts of different scales.

\end{abstract}

\section{Introduction}
\label{sec:introduction}

Skin lesion segmentation and classification are two core tasks in computer-aided diagnosis (CAD) of dermoscopic images. Both are clinically consequential, particularly for malignant melanoma, where early detection is critical for patient survival \cite{ernst2022melanoma,schadendorf2018melanoma,siegel2023cancer}. Modern dermoscopic CAD is dominated by deep learning \cite{kreouzi2024deep}. Convolutional networks capture fine-grained local textures while Vision Transformers (ViTs) model long-range dependencies \cite{kwiatkowska2021cnn, dosovitskiy2020image, garcia2025clinical}. Both signal types carry diagnostic weight: a single pigment globule and the lesion-wide morphology both inform the diagnosis. CNN–Transformer hybrid architectures have therefore become a natural fit for this domain \cite{tanha2025lighthybridvit, sm2025hybridcnntransformer, hasan2025hybridcnnvit}.

Research on jointly modeling the two tasks has followed two parallel lines, each addressing one dimension of the problem while leaving the other unattended. A first line addresses architectural coupling through end-to-end multi-task frameworks that share a single encoder across both tasks. Xie et al. fed a predicted coarse mask into the classification branch as an auxiliary input, and Zhang et al. introduced a contrastive objective that aligns classification and segmentation features inside the lesion region \cite{Xie2020MutualBootstrapping, Zhang2025MTCLSkin}. These approaches couple the two pathways at the architectural level, but the clinical reasoning that dermatologists use to interpret a lesion is not encoded into the model. A second line addresses clinical prior injection, most often through the ABCD rule (Asymmetry, Border, Color, Diameter). Guo et al. and Kotla et al. incorporated this rule through attention and auxiliary supervision respectively \cite{guo2025,Kotla2025MultiTaskSkinLesion}. As Bigliardi and Karels et al. point out, however, the ABCD rule 
was developed for naked-eye assessment of macroscopic features. 
It was not designed for the microscopic structural cues that 
dermatologists actually attend to when reading dermoscopic 
images \cite{Bigliardi2025LB1046, Karels2026PatchTest3DImaging}.

Pattern Analysis is the diagnostic framework dermatologists apply when reading dermoscopic images. It bases the diagnosis on a defined set of microscopic structures: the pigment network, streaks, dots and globules, regression areas, and the blue-whitish veil \cite{Carli2003PatternAnalysis}. Although small-scale datasets such as PH2 provide structure-level annotations, the large-scale training corpora used in dermoscopic deep learning (such as International Skin Imaging Collaboration (ISIC)) do not, which has limited direct adoption of Pattern Analysis in the deep learning literature. An underexplored alternative is to design architectural inductive biases inspired by the visual cues that Pattern Analysis attends to, rather than to supervise the network to detect canonical structures explicitly. We consider three such cue families: abrupt local chromatic transitions, scale-diverse pigment structures, and internal structural asymmetry. Each can motivate an architectural component without requiring structure-level supervision.

Building on this observation, we propose the Multi-task Dermoscopic Skin-lesion Network (MDSkin-Net), which couples the two tasks end-to-end within a single framework. Our contributions are summarized as follows:
\begin{itemize}

\item \textbf{Hybrid CNN-Transformer shared encoder.} We adopt a ConvNeXt-Tiny stem coupled with a MobileViT bottleneck as the shared encoder, which preserves the local inductive biases needed for fine-grained dermoscopic textures while modeling lesion-wide global dependencies.

\item \textbf{Pattern Analysis-Guided Attention Module (PAGAM).} We introduce three architectural priors informed by the three cue families identified above: an improved Efficient Channel Attention (iECA) designed to be sensitive to abrupt chromatic variations; a Multi-Scale Spatial Attention (MSSA) whose receptive fields are matched to the scale diversity of pigment structures; and a Biased Asymmetry Attention (BAA) that injects a patch-pair geometric prior into the MobileViT bottleneck self-attention.

\item \textbf{Multi-Scale Spatial Alignment Regularization (MSAR).} We further introduce a hierarchical alignment objective that uses the segmentation ground-truth mask as soft spatial supervision, confining the classification head to lesion-localized evidence and aligning both task pathways under a shared spatial prior.

\end{itemize}

Trained exclusively on the official ISIC 2017 training set without external dermoscopy data, MDSkin-Net transfers consistently to PH2 and ISIC 2018 under zero-shot evaluation, with in-domain classification competitive and segmentation trading peak accuracy for cross-cohort robustness.

\section{Methodology}

Figure 1 illustrates the overall architecture of the proposed \textbf{MDSkin-Net}. Our framework follows a U-Net-style dual-branch topology for joint segmentation and classification \cite{ronneberger2015unet}.

\subsection{Hybrid CNN-Transformer Shared Encoder}
\label{subsec:shared}

We adopt a ConvNeXt-Tiny backbone~\cite{liu2022convnext} for hierarchical feature extraction, balancing computational efficiency and representational capacity. The backbone comprises four downsampling stages with block depths $\{3, 3, 9, 3\}$ and channel dimensions $C \in \{96, 192, 384, 768\}$. To preserve fine spatial details and ensure compatibility with the segmentation decoder, depthwise--pointwise convolutions activated by Gaussian Error Linear Units (GELU) reduce these channels to $\{64, 128, 256, 512\}$.

The deepest feature map from the fourth stage serves as the input to the bottleneck MobileViT block~\cite{mehta2021mobilevit}. A $1 \times 1$ pointwise convolution first compresses the channel dimension from $512$ to $d = 256$ to reduce computational cost. Let $X_{\text{in}} \in \mathbb{R}^{d \times H' \times W'}$ denote this compressed tensor. To capture both local detail and global context, the block applies the following hybrid transformation:

\begin{equation}
\begin{aligned}
    F_{local}  &= \text{Conv}_{1\times1}\bigl(\text{DWConv}_{3\times3}(X_{in})\bigr), \\
    Z_U        &= \mathcal{P}(F_{local}), \\
    \hat{Z}_A  &= \text{MHSA}\bigl(\text{RMSNorm}(Z_U)\bigr) + Z_U, \\
    Z_T        &= \text{FFN}\bigl(\text{RMSNorm}(\hat{Z}_A)\bigr) + \hat{Z}_A, \\
    F_{global} &= \text{Conv}_{1\times1}\bigl(\mathcal{P}^{-1}(Z_T)\bigr), \\
    X_{out}    &= X_{in} + \text{Conv}_{1\times1}\bigl(\text{Concat}[X_{in}, F_{global}]\bigr).
\end{aligned}
\end{equation}

Here, $\text{DWConv}_{3\times3}$ and $\text{Conv}_{1\times1}$ denote the depthwise and pointwise convolutions. The unfold operation $\mathcal{P}(\cdot)$ reshapes the spatial tensor into $N$ non-overlapping patches of $P$ pixels each, stacked as $Z_U \in \mathbb{R}^{P \times N \times d}$.

Unlike standard ViT attention, which flattens all patch tokens within a global sequence, the Multi-Head Self-Attention (MHSA) here operates across patches at each fixed pixel position. For each relative pixel position $p \in \{1, \dots, P\}$, attention is computed across the $N$ patches:
\begin{equation}
\label{eq:multi-head}
\begin{aligned}
    \hat{Z}_A^{(p)} = \text{Softmax}\left(\frac{Q^{(p)} (K^{(p)})^T}{\sqrt{d_k}}\right) V^{(p)}
\end{aligned}
\end{equation}

where $Q^{(p)}$, $K^{(p)}$ and $V^{(p)}$ denote the query, key and value matrices, each obtained by linearly projecting the $p$-th pixel over all $N$ patches, and $d_k$ scales the dot product. $\text{RMSNorm}(\cdot)$ denotes Root Mean Square Normalization, adopted for its computational efficiency, while $\text{FFN}(\cdot)$ is the position-wise feed-forward network. 

The inverse operation $\mathcal{P}^{-1}(\cdot)$ folds the tokens back to the spatial layout, and a subsequent $1 \times 1$ convolution yields the global representation $F_{\text{global}}$. This tensor is then concatenated with $X_{\text{in}}$ along the channel dimension, passed through a final pointwise convolution, and added back to $X_{\text{in}}$ as a residual connection. 

Notably, the MHSA in Eq.~\ref{eq:multi-head} serves as the injection point for our Biased Asymmetry Attention (Section~\ref{subsection:BAA}), which augments standard self-attention with an internal structural asymmetry prior.

\subsection{Pattern Analysis-Guided Attention Module (PAGAM)}
PAGAM is not intended to detect canonical dermoscopic structures explicitly. Rather, it embeds architectural priors at the cue level, aligned with three categories of visual cues central to Pattern Analysis.

\subsubsection{Chromatic-Sensitive Channel Attention (iECA)}
Motivated by abrupt chromatic transitions such as those in regression areas and the blue-whitish veil, the iECA module augments standard ECA with a spatial standard-deviation descriptor~\cite{wang2020eca}.

\begin{figure*}[t]
    \centering
    \includegraphics[trim={0cm 13cm 2cm 3.6cm}, clip, width=0.85\textwidth]{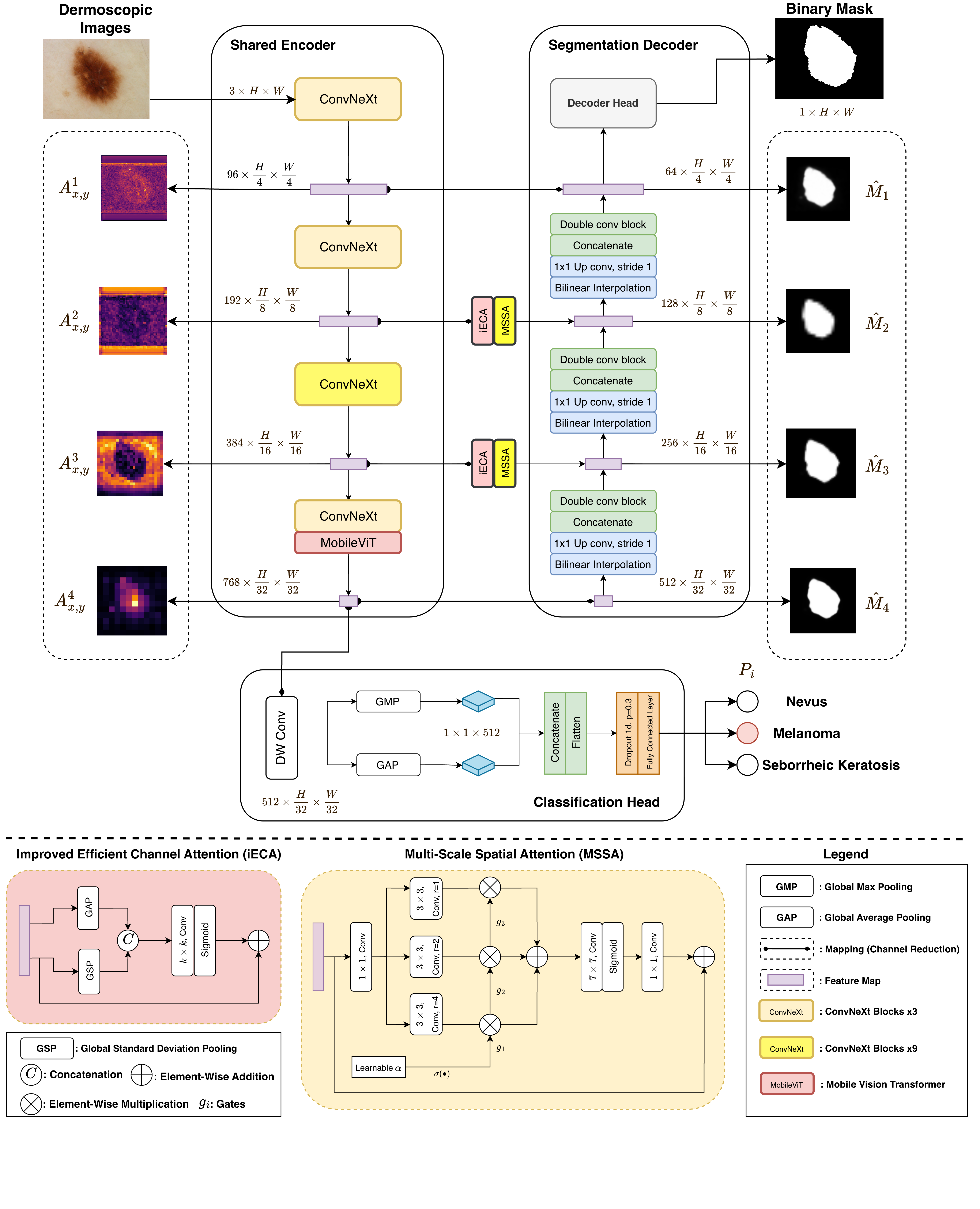}
    \captionsetup{justification=centering}
    \caption{Architecture of the Proposed MDSkin-Net}
    \label{fig:main}
\end{figure*}

Given an input tensor $X$ of size $C \times H \times W$, iECA pools two statistics along the spatial axes in parallel, the channel-wise mean and the channel-wise standard deviation:
\begin{equation}
\begin{aligned}
    F_{avg} &= \text{AvgPool}(X) \in \mathbb{R}^{C}, \\
    F_{std} &= \text{StdPool}(X) \in \mathbb{R}^{C}.
\end{aligned}
\end{equation}

Instead of a linear projection, these two statistical descriptors are reshaped and concatenated to form a dual-descriptor tensor $F_{concat} \in \mathbb{R}^{2 \times C}$. We then directly fuse these descriptors using a 1D convolution over the channel dimension:
\begin{equation}
\begin{aligned}
    A_{sem} &= \sigma(\text{Conv1D}_{k}(F_{concat})), \\
    \hat{X}_{sem} &= A_{sem} \otimes X.
\end{aligned}
\end{equation}

Here, $\text{Conv1D}_{k}$ applies a 1D convolution with a dynamically determined kernel size $k$ and $2$ input channels. This operation captures cross-channel interactions while merging the mean and variance statistics. The sigmoid $\sigma(\cdot)$ then turns the fused response into channel attention weights $A_{sem} \in \mathbb{R}^{C \times 1 \times 1}$; these rescale $X$ element-wise ($\otimes$) into the semantically calibrated representation $\hat{X}_{sem}$.

\subsubsection{Multi-Scale Spatial Attention (MSSA)}
Pigment structures span a wide scale range, from localized dots and globules to the lesion-wide pigment network. The MSSA module aggregates features across receptive fields matched to this range.

Take an input $X$ of size $C_1 \times H \times W$. Its channels are first compressed to $C_2$ by a $1 \times 1$ convolution, and the resulting $X_{\mathrm{red}}$ passes through three parallel $3 \times 3$ depthwise convolutions at dilation rates $r \in \{1, 2, 4\}$, which target progressively larger structures, from dots ($r = 1$) to the pigment network ($r = 4$). Instead of unweighted addition, we introduce learnable gating parameters $\alpha \in \mathbb{R}^3$, activated by a sigmoid function $\sigma(\cdot)$, to dynamically aggregate these diverse receptive fields:
\begin{equation}
    M = \text{GELU}\left(\text{GN}\left(X_{red} + \sum_{i=1}^{3} \sigma(\alpha_i) \text{DWConv}_{3\times3}^{(r_i)}(X_{red})\right)\right)
\end{equation}
where $\text{GN}$ denotes Group Normalization, yielding the multi-scale aggregated feature $M \in \mathbb{R}^{C_2 \times H \times W}$.

To build the spatial saliency mask from these multi-scale patterns, $M$ undergoes channel-wise average and max pooling. A $7 \times 7$ convolution then maps the concatenated descriptors to the attention mask $A_{spatial}$. This mask is applied to $M$, expanded back to $C_1$ channels via pointwise convolution, and fused with the original input $X$ via a residual connection to ensure optimization stability:
\begin{equation}
\begin{aligned}
    A_{spatial} &= \sigma(\text{Conv}_{7\times7}(\text{Concat}[\text{AvgPool}_{c}(M), \text{MaxPool}_{c}(M)])), \\
    X_{out} &= \text{GELU}(\text{GN}(\text{Conv}_{1\times1}(M \otimes A_{spatial})) + X).
\end{aligned}
\end{equation}

\subsubsection{Asymmetry-Aware Self-Attention (BAA)}
\label{subsection:BAA}
Asymmetry is a Pattern Analysis cue that differs from the lesion-shape asymmetry of the ABCD rule. Whereas the ABCD rule evaluates whether the overall lesion contour is bilaterally symmetric at the macroscopic level, dermoscopic Pattern Analysis assesses asymmetry at a patch-level granularity defined by structural deviations between internal regions of the lesion. A lesion with a roughly circular contour may still be considered asymmetric under this view if one half exhibits a typical pigment network while the other half presents structureless areas. Motivated by this distinction, we propose Biased Asymmetry Attention (BAA), which introduces an explicit bias term into the self-attention operation, designed to encourage attention toward patch-level structural asymmetry.

Because lesions are rarely centered at the geometric image center, BAA first estimates a data-driven reference point weighted by feature saliency, around which the asymmetry-related bias is constructed. Let $X \in \mathbb{R}^{B \cdot P \times N \times D}$ denote the input feature tensor entering the attention block, where the batch and $P$ pixel positions are folded into the leading axis so that attention runs independently for each $(b,p)$ pair over the $N$ patches, as in Eq.~(2). Let $Q, K, V \in \mathbb{R}^{B \cdot P \times H \times N \times d_k}$ be the query, key, and value tensors with per-head dimension $d_k = D/H$, and let $\mathbf{u}_n \in \mathbb{R}^2$ denote the 2D coordinate of the $n$-th patch. Treating the $\ell_2$ norm of $X$ (averaged over the pixel dimension) as saliency, we derive the saliency-weighted spatial centroid $\mathbf{c}_b \in \mathbb{R}^2$ for the $b$-th image as
\begin{equation}
    \mathbf{c}_b = \sum_{n=1}^{N} \left( \frac{\mathbb{E}_{p} \left[ \| X_{b, p, n} \|_2 \right]}{\sum_{m=1}^{N} \mathbb{E}_{p} \left[ \| X_{b, p, m} \|_2 \right] + \epsilon} \right) \mathbf{u}_n,
\end{equation}
where $\epsilon$ prevents division by zero and $\mathbb{E}_{p}[\cdot]$ averages over the $P$ pixel positions, reducing the pixel dimension. With $\mathbf{c}_b$ serving as the lesion-aware reference point, a geometric component is introduced to softly weight patch pairs according to their spatial layout relative to $\mathbf{c}_b$, through a Gaussian kernel whose bandwidth $\sigma_h$ is a learnable scalar parameter indexed by the head dimension $h \in \{1, \ldots, H\}$, which allows each head to adopt its own spatial tolerance:
\begin{equation}
    \mathcal{K}_{i,j}^{geom} = \exp\left( - \frac{\|\mathbf{u}_i + \mathbf{u}_j - 2\mathbf{c}_b\|_2^2}{2\sigma_h^2} \right) \in \mathbb{R}^{B \times H \times N \times N}.
\end{equation}
The geometric kernel alone reflects only spatial layout; we therefore complement it with a feature-level term intended to reflect the structural discrepancy between patches. Consistent with the pixel-dimension reduction above, each patch is represented by its pixel-averaged token $x_i = \mathbb{E}_{p}\left[ X_{b,p,i} \right] \in \mathbb{R}^D$, and the discrepancy is taken as the squared $\ell_2$ distance:
\begin{equation}
    \mathcal{D}_{i,j}^{feat} = \|x_i - x_j\|_2^2 = \|x_i\|_2^2 + \|x_j\|_2^2 - 2 \langle x_i, x_j \rangle \in \mathbb{R}^{B \times N \times N}.
\end{equation}
The two terms are coupled via Hadamard product and subsequently normalized by their per-sample, per-head maximum for numerical stability:
\begin{equation}
    \hat{B}_{i,j} = \frac{\mathcal{D}_{i,j}^{feat} \odot \mathcal{K}_{i,j}^{geom}}{\max_{i,j}(\mathcal{D}_{i,j}^{feat} \odot \mathcal{K}_{i,j}^{geom}) + \epsilon} \in \mathbb{R}^{B \times H \times N \times N}.
\end{equation}
Finally, the resulting bias $\hat{B}$ is injected into multi-head self-attention through a learnable head-wise scaling factor $\Gamma \in \mathbb{R}^{H}$, which lets each head adjust how strongly the bias modulates its attention. Since $\hat{B}$ is defined at the patch-pair level and carries no pixel index, it is broadcast along the pixel dimension and added identically at each position $p$. Denoting the biased logits at pixel position $p$ by
\begin{equation}
    S^{(p)} = \frac{Q^{(p)}(K^{(p)})^{\top}}{\sqrt{d_k}} + \Gamma \odot \hat{B},
\end{equation}
the per-pixel BAA output is
\begin{equation}
\begin{split}
\mathrm{Attention}^{(p)}(Q, K, V)
&= \mathrm{Softmax}\!\left( S^{(p)} \right) V^{(p)}, \\
&\quad \forall\, p \in \{1, \ldots, P\}.
\end{split}
\end{equation}

\subsection{Multi-Task Decoding Heads}
To fulfill the dual-task prediction efficiently, the network branches into a progressive segmentation decoder and a spatial-decoupled classification head.

\subsubsection{Segmentation Decoder} 
The segmentation branch progressively restores spatial resolution while fusing hierarchical semantics. To reconcile the semantic disparity between deep and shallow features, the encoder skip-connection feature $E_i$ at stage $i$ is conditionally modulated before fusion. Let $\tilde{E}_i$ denote the modulated skip-connection:
\begin{equation}
    \tilde{E}_i = \begin{cases} 
    \text{MSSA}(\text{iECA}(E_i)), & \text{if } i \in \{2, 3\} \\
    E_i, & \text{otherwise} 
    \end{cases}
\label{eq:skip_modulation}
\end{equation}
Subsequently, the stage-wise decoder feature $D_i$ is recursively computed by bilinearly upsampling $D_{i+1}$, refining the result with a lightweight ECA channel-attention block, halving its channels via a $1 \times 1$ convolution, concatenating with $\tilde{E}_i$ along the channel dimension ($\parallel$), and applying a composite transformation block $\mathcal{F}$:
\begin{equation}
    D_i = \mathcal{F}\!\left( \text{Conv}_{1\times 1}\!\left(\text{ECA}(\text{Up}(D_{i+1}))\right) \parallel \tilde{E}_i \right)
\label{eq:decoder}
\end{equation}
where $\mathcal{F}(\cdot)$ stacks two $\text{Conv}_{3\times 3}$--$\text{GN}$--$\text{GELU}$ blocks for feature transformation. 

\begin{table}[!t]
\centering
\caption{\textbf{Dermoscopy cohorts used in this work.} Only ISIC 2017 contributes training data; the other two are held out entirely and seen only at inference.}
\label{tab:datasets}
\renewcommand{\arraystretch}{1.15}
\setlength{\tabcolsep}{3pt}
\footnotesize
\begin{tabular*}{\columnwidth}{@{\extracolsep{\fill}}lccl@{}}
\toprule
Cohort & Images & Annotation & Role here \\
\midrule
ISIC 2017~\cite{codella2017isic}
  & 2{,}750 & label + mask & train / val / test \\
PH2~\cite{mendonca2013ph2}
  & 200 & label + mask & zero-shot, small \\
ISIC 2018 T1~\cite{codella2018isic2018, tschandl2018ham10000}
  & 1{,}000 & mask only & zero-shot, large \\
\bottomrule
\end{tabular*}
\\[3pt]
\raggedright
\footnotesize
ISIC 2017 composition: 1{,}843 nevus, 386 seborrheic keratosis, 521 melanoma; official split 2{,}000 / 150 / 600.
PH2 composition: 80 common nevi, 80 atypical nevi, 40 melanomas.
ISIC 2018 Task~1 carries no image-level diagnosis, so it scores segmentation only.
\end{table}

\subsubsection{Classification Head} 
Concurrently, the bottleneck embedding $X_{bot}$ is routed to an independent classification head. We sequentially apply depthwise ($dw$) and pointwise ($pw$) convolutions to disentangle spatial and cross-channel correlations. Leveraging the previously defined composite block $\mathcal{F}$, the decoupled feature is obtained as $X_{cls} = \mathcal{F}_{pw}(\mathcal{F}_{dw}(X_{bot}))$.

Then, $X_{cls}$ is aggregated by a dual-pooling strategy. The decoupled feature is summarized by two global poolings, average ($\text{GAP}$) and max ($\text{GMP}$), concatenated along the channel axis and passed to a two-layer MLP with GELU activation and dropout, which yields the final diagnostic prediction $\hat{Q}$:
\begin{equation}
    \hat{Q} = \text{MLP}\left( \text{GAP}(X_{cls}) \parallel \text{GMP}(X_{cls}) \right)
\end{equation}

\subsection{Optimization and Loss Formulation}
To jointly optimize the proposed MDSkin-Net, the overall objective function is formulated as a combination of three distinct components.

\subsubsection{Segmentation Loss}We use a composite of BCE and Dice loss \cite{sudre2017generalised}:
\begin{equation}
    \mathcal{L}_{base}(\hat{M}, M) = 0.5 \cdot \mathcal{L}_{BCE}(\hat{M}, M) + 0.5 \cdot \mathcal{L}_{Dice}(\hat{M}, M)
\end{equation}

In addition, deep supervision is added at each decoder stage \cite{li2019deepsupervision}. Let $\{\hat{M}_i\}_{i=0}^{N}$ denote the multi-scale predictions, where $\hat{M}_0$ is the final full-resolution mask and $\hat{M}_{i \ge 1}$ are intermediate auxiliary outputs. The total segmentation objective is formulated as:
\begin{equation}
    \mathcal{L}_{seg} = \sum_{i=0}^{N} 0.5^i \cdot \mathcal{L}_{base}(\hat{M}_i, M)
\end{equation}

\subsubsection{Classification Loss}
The diagnostic classification branch is optimized using a label-smoothed cross-entropy (CE) loss \cite{guo2024cross}. Let $Q$
denote the one-hot ground-truth label and $\hat{Q}$ the diagnostic logits over $K$ categories. With smoothing factor $\epsilon_{ls}$, the soft target is defined as

\begin{equation}
\tilde{Q}^{(k)} = (1 - \epsilon_{ls}) Q^{(k)} + \frac{\epsilon_{ls}}{K},
\end{equation}

and the classification objective is given by
\begin{equation}
\mathcal{L}_{\mathrm{cls}} = - \sum_{k=1}^{K} \tilde{Q}^{(k)} \log \left( \hat{Q}^{(k)} \right).
\end{equation}

\begin{table*}[!t]
\centering
\renewcommand{\arraystretch}{1.0}
\setlength{\tabcolsep}{3.5pt}
\footnotesize
\caption{\textbf{Comparison with state-of-the-art methods on skin lesion benchmarks.}
(a) Cross-dataset generalization on the PH2 dataset (small cohort) under zero-shot evaluation;
(b) Segmentation on the ISIC2017 official test set;
(c) Classification on the ISIC2017 official test set.}
\label{tab:sota_combined}

\centerline{\textbf{(a) PH2 cross-dataset generalization (zero-shot, multi-task)}}
\vspace{2pt}
\label{tab:sota_ph2_zeroshot}
\begin{tabular*}{\textwidth}{@{\extracolsep{\fill}}lc|cccc|cccc@{}}
\toprule
\multirow{2}{*}{Method} & \multirow{2}{*}{Year}
& \multicolumn{4}{c|}{Segmentation}
& \multicolumn{4}{c}{Classification} \\
\cmidrule(lr){3-6} \cmidrule(lr){7-10}
& & DSC & SE & SP & ACC & AUC & SE & SP & ACC \\
\midrule
U-Net~\cite{ronneberger2015unet}           & 2015 & 89.36 & 91.25 & 95.88 & 92.33 & — & — & — & — \\
PSPNet~\cite{pspnet}                       & 2017 & 84.55 & 79.82 & 93.65 & 86.33 & — & — & — & — \\
Attn-UNet~\cite{oktay2018attention}        & 2018 & 90.03 & 92.05 & \textbf{96.40} & 92.76 & — & — & — & — \\
UNet++~\cite{zhou2018unetpp}               & 2018 & 88.58 & 88.20 & 92.44 & 90.31 & — & — & — & — \\
MB-DCNN$^{\dagger}$~\cite{Xie2020MutualBootstrapping}   & 2020 & \textbf{92.60} & 97.90 & 95.30 & \textbf{96.50} & 95.60 & \textbf{82.50} & 90.00 & 88.50 \\
DoubleU-Net~\cite{doubleunet}              & 2020 & 80.55 & 78.94 & 90.49 & 84.60 & — & — & — & — \\
MCG-Net~\cite{asadi2020mcgunet}                    & 2020 & 79.06 & 89.81 & 92.95 & 83.53 & — & — & — & — \\
TransUNet~\cite{chen2021transunet}        & 2021 & 88.40 & 90.63 & 94.27 & 92.00 & — & — & — & — \\
MALUNet~\cite{ruan2022malunet}             & 2022 & 85.03 & 85.41 & 94.72 & 90.19 & — & — & — & — \\
TransNorm~\cite{transnorm}                 & 2022 & 85.92 & 92.33 & 94.64 & 89.74 & — & — & — & — \\
DHU-Net~\cite{dhunet}                      & 2023 & 87.59 & 91.28 & 95.98 & 92.25 & — & — & — & — \\
HSH-Unet~\cite{wu2024hshunet}                      & 2023 & 92.26 & 91.59 & 96.69 & 95.04 & — & — & — & — \\
SVMB-Net~\cite{svmbnet}                    & 2025 & 91.86 & 94.53 & 95.34 & 95.37 & — & — & — & — \\
\midrule
\textbf{Ours}                              & 2026 & 92.28 & 98.07 & 96.21 & 96.38 & 95.38 & 75.00 & 97.08 & 92.67 \\
\textbf{Ours (ens.)}                       & 2026 & 92.38 & \textbf{98.14} & 96.27 & 96.43 & \textbf{97.84} & 80.00 & \textbf{98.12} & \textbf{94.50} \\
\bottomrule
\end{tabular*}

\vspace{12pt}

\centerline{\textbf{(b) ISIC2017 segmentation}}
\vspace{2pt}
\label{tab:sota_isic2017_seg}
\begin{tabular*}{\textwidth}{@{\extracolsep{\fill}}lccccccc@{}}
\toprule
Method & Year & Params(M) & FLOPs(G) & DSC & IoU & SE & SP \\
\midrule
U-Net~\cite{ronneberger2015unet}              & 2015 & 31.04   & 123.16 & 81.59 & 72.34 & 81.72 & 96.80 \\
Att U-Net~\cite{oktay2018attention}           & 2018 & 34.88   & 149.92 & 80.82 & 71.73 & 79.98 & 97.76 \\
U-Net++~\cite{zhou2018unetpp}                 & 2018 & 9.16    & 78.53  & 88.58 & 70.00 & 87.20 & 92.44 \\
MB-DCNN$^{\dagger}$~\cite{Xie2020MutualBootstrapping}      & 2020 & 130.29  & —      & 87.80 & 80.40 & 87.40 & 96.80 \\
TransUNet~\cite{chen2021transunet}            & 2021 & 105.57  & 72.59  & 81.23 & 71.47 & 82.63 & 95.77 \\
MALUNet$^{\star}$~\cite{ruan2022malunet}      & 2022 & 0.18    & 0.19   & 88.13 & 78.78 & 84.78 & 98.47 \\
SwinUNet~\cite{cao2022swinunet}               & 2022 & 27.22   & 17.43  & \textbf{91.83} & —     & \textbf{91.42} & 97.98 \\
VM-UNet$^{\star}$~\cite{ruan2024vmunet}       & 2024 & 44.27   & —      & 89.03 & 80.23 & 89.90 & 97.58 \\
CA Y-Net$^{\dagger}$~\cite{Zhang2025MTCLSkin}           & 2025 & —       & —      & 87.30 & 79.60 & —     & —     \\
SVMB-Net~\cite{svmbnet}                       & 2025 & 45.57   & 163.70 & 90.56 & \textbf{83.45} & 92.10 & 98.48 \\
\midrule
\textbf{Ours}                                 & 2026 & 34.06 & 18.59 & 84.63 & 76.16 & 80.05 & 99.36 \\
\textbf{Ours (ens.)}                          & 2026 & 34.06 & 18.59 & 84.72 & 76.29 & 80.08 & \textbf{99.38} \\
\bottomrule
\end{tabular*}

\vspace{12pt}

\centerline{\textbf{(c) ISIC2017 classification}}
\vspace{2pt}
\label{tab:sota_isic2017_cls}
\begin{tabular*}{\textwidth}{@{\extracolsep{\fill}}lc|ccc|ccc|c@{}}
\toprule
\multirow{2}{*}{Method} & \multirow{2}{*}{Year}
& \multicolumn{3}{c|}{Melanoma}
& \multicolumn{3}{c|}{Seborrheic Keratosis}
& \multirow{2}{*}{\shortstack{Mean\\AUC}} \\
\cmidrule(lr){3-5} \cmidrule(lr){6-8}
& & AUC & SE & SP & AUC & SE & SP & \\
\midrule
\#1 ISIC2017 leaderboard          & 2017 & 86.8 & \textbf{73.5} & 85.1          & 95.3          & \textbf{97.8} & 77.3          & 91.1 \\
\#2 ISIC2017 leaderboard          & 2017 & 85.6 & 10.3          & \textbf{99.8} & 96.5          & 17.8          & \textbf{99.8} & 91.0 \\
\#3 ISIC2017 leaderboard          & 2017 & 87.4 & 54.7          & 95.0          & 94.3          & 35.6          & 99.0          & 90.8 \\
\#4 ISIC2017 leaderboard          & 2017 & 87.0 & 42.7          & 96.3          & 92.1          & 58.9          & 97.6          & 89.6 \\
\#5 ISIC2017 leaderboard          & 2017 & 83.0 & 43.6          & 92.5          & 94.2          & 70.0          & 99.5          & 88.6 \\
ARL-CNN~\cite{arlcnn}             & 2019 & 87.5 & 65.8          & 89.6          & 95.8          & 87.8          & 86.7          & 91.7 \\
MB-DCNN$^{\dagger}$~\cite{Xie2020MutualBootstrapping}  & 2020 & 90.3          & 72.7 & 91.5          & \textbf{97.3} & 84.4 & 94.5 & \textbf{93.8} \\
CA Y-Net$^{\dagger}$~\cite{Zhang2025MTCLSkin}       & 2025 & \textbf{90.4} & 66.7 & 90.1          & 95.3          & 77.8 & 94.1 & 92.8 \\
\midrule
\textbf{Ours}                     & 2026 & 87.3 & 53.0 & 94.6 & 94.9 & 83.7 & 93.4 & 91.1 \\
\textbf{Ours (ens.)}              & 2026 & 88.3 & 52.1 & 94.8 & 94.9 & 84.4 & 93.3 & 91.6 \\
\bottomrule
\end{tabular*}

\vspace{3pt}
\raggedright
\scriptsize
$^{\dagger}$~Trained with additional ISIC archive images (more than one thousand) beyond the official ISIC2017 training set.
$^{\star}$~Results obtained on a 7:3 random split of ISIC2017 (not the official test set).
``—'' indicates the metric is not reported in the original source. 
\textbf{Bold}: best result across all listed methods. FLOPs computed at $384\times384$ input.
\textbf{Ours}: per-seed metrics averaged over three independent training runs (seeds: 42, 1024, 2024).
\textbf{Ours (ens.)}: ensemble of the three runs, obtained by averaging predicted logits before computing metrics.
\end{table*}

\subsubsection{Multi-Scale Spatial Alignment Regularization (MSAR)}
\label{sec:align}
Since dense prediction and image-level labeling operate at different granularities, we introduce MSAR ($\mathcal{L}_{align}$), which comprises a residual self-attentive spatial gate and an auxiliary Binary Cross-Entropy (BCE) supervision while providing hierarchical lesion-aligned guidance to the decoder. The deepest gate ($l = 4$) is shared between the classification head and the segmentation decoder, thereby coupling the two tasks; the shallower gates ($l \in \{1, 2, 3\}$) impose a coarse-to-fine lesion-localization prior on the segmentation pathway.

At each encoder stage $l \in \{1, \ldots, L\}$, a $1 \times 1$ convolution projects the skip feature $\tilde{E}_l$ (Eq.~\eqref{eq:skip_modulation}) into a single-channel activation map $A^{(l)} \in \mathbb{R}^{H_l \times W_l}$. The sigmoid-activated map $\sigma(A^{(l)})$ defines a soft spatial gate that residually modulates $\tilde{E}_l$:
\begin{equation}
    \tilde{E}^{gate}_l = \tilde{E}_l \odot \big(1 + \alpha_l \cdot \sigma(A^{(l)})\big),
    \label{eq:align_gate}
\end{equation}
where $\alpha_l \in \mathbb{R}$ is a learnable per-stage scalar (initialized to $0.1$) and the residual form preserves $\tilde{E}_l$ as a baseline. The gated feature $\tilde{E}^{gate}_l$ replaces $\tilde{E}_l$ as the skip-connection input in Eq.~\eqref{eq:decoder} for $l \in \{1, 2, 3\}$, and as the input to the MobileViT bottleneck (and thus the classification head) for $l = 4$.

To drive each gate toward pathologically meaningful regions, we supervise $A^{(l)}$ with a multi-scale BCE objective. The ground-truth mask $M \in \{0,1\}^{H \times W}$ is downsampled via adaptive average pooling rather than nearest-neighbor interpolation, preserving partial occupancy near lesion boundaries:
\begin{equation}
    M^{(l)} = \text{AdaptiveAvgPool}(M, (H_l, W_l)) \in [0,1]^{H_l \times W_l}.
\end{equation}
The per-stage alignment term is the standard BCE between $A^{(l)}$ and $M^{(l)}$:
\begin{equation}
    \begin{aligned}
    \mathcal{L}_{align}^{(l)} = -\frac{1}{H_l W_l} \sum_{x,y} \big[ &M^{(l)}_{x,y} \log \sigma(A^{(l)}_{x,y}) \\
    &+ (1 - M^{(l)}_{x,y}) \log(1 - \sigma(A^{(l)}_{x,y})) \big].
    \end{aligned}
\end{equation}
The total alignment loss aggregates the per-stage terms with normalized weights $\{w_l\}_{l=1}^{L}$, where larger weights are assigned to deeper stages:
\begin{equation}
    \mathcal{L}_{align} = \sum_{l=1}^{L} \frac{w_l}{\sum_{k=1}^{L} w_k} \cdot \mathcal{L}_{align}^{(l)}.
\end{equation}
The overall objective combines all three task losses:
\begin{equation}
    \mathcal{L}_{total} = \omega_{seg}\mathcal{L}_{seg} + \omega_{cls} \mathcal{L}_{cls} + \omega_{align}\mathcal{L}_{align},
\end{equation}
with scalar coefficients $\omega_{seg}$, $\omega_{cls}$, $\omega_{align}$ and per-stage weights $\{w_l\}$ given in Section~\ref{subsection:hyperparameters}.

\begin{table}[t]
\centering
\caption{\textbf{Cross-dataset generalization on the ISIC 2018 dataset (larger cohort) under zero-shot segmentation evaluation.}}
\label{tab:isic2018_zeroshot}
\renewcommand{\arraystretch}{1.15}
\setlength{\tabcolsep}{1pt} 
\footnotesize
\begin{tabular*}{\columnwidth}{@{\extracolsep{\fill}}lccccc@{}}
\toprule
Method & DSC & IoU & SE & SP & ACC \\
\midrule
Ours & $88.81{\scriptstyle\pm 0.45}$ & $81.30{\scriptstyle\pm 0.69}$ & $94.31{\scriptstyle\pm 0.10}$ & $96.48{\scriptstyle\pm 0.31}$ & $95.38{\scriptstyle\pm 0.21}$ \\
\textbf{Ours (ens.)} & \textbf{88.92} & \textbf{81.45} & \textbf{94.39} & \textbf{96.51} & \textbf{95.42} \\
\bottomrule
\end{tabular*}
\\[3pt]
\raggedright
\footnotesize
Results are obtained by predictions from three independent training runs (seeds: 42, 1024, 2024) on $n=1000$ images.
\end{table}

\section{EXPERIMENTS}
\subsection{Datasets}
Three public dermoscopy datasets are used in this work. Their composition and role are summarized in Table~\ref{tab:datasets}.

We train MDSkin-Net solely on the ISIC 2017 training set, following the ISIC 2017 challenge protocol. Classification is formulated as two binary tasks, melanoma versus the remaining classes and seborrheic keratosis versus the remaining classes, with the mean of the two tasks reported as the overall classification performance. On PH2, classification is evaluated as a single binary task (melanoma versus non-melanoma), consistent with the dataset's label granularity. The ISIC 2018 Task~1 test set carries no image-level diagnostic labels and is therefore used for segmentation evaluation only. PH2 and ISIC 2018 are used exclusively as external benchmarks under zero-shot evaluation, without any fine-tuning, to assess robustness and cross-dataset generalization. Unlike prior multi-task and related approaches \cite{Zhang2025MTCLSkin, Xie2020MutualBootstrapping, arlcnn, matsunaga2017, gonzalez2019dermaknet, bi2017, zhang2019sdl, xie2019semi, chen2018mdnn}, all experiments are conducted strictly on the official ISIC 2017 partition without incorporating any external dermoscopy data or random splits.

\subsection{Data Augmentation}
The original training set is class-imbalanced, and no external dermoscopy data were added to offset this; instead, a WeightedRandomSampler rebalanced the sampling distribution. Each image was scaled to a $384 \times 384$ resolution, with every spatial transform applied jointly to the image and its mask so that pixel-level correspondence was retained. The spatial transforms were horizontal and vertical flipping, rotation, and elastic deformation; the photometric transforms were gamma correction, Gaussian blurring, color jittering, and contrast-limited adaptive histogram equalization (CLAHE). Inputs were finally standardized with ImageNet statistics.

Multi-scale test-time augmentation (TTA) was used at inference. Every image was evaluated at three resolutions ($256 \times 256$, $384 \times 384$, $512 \times 512$) and under the four flip states obtained from combining horizontal and vertical flips, producing twelve forward passes whose classification and segmentation logits were then averaged.

\subsection{Experimental Setup}
\subsubsection{Hyper-Parameters Configuration}
\label{subsection:hyperparameters}
Table~\ref{tab:hyperparams} summarizes the training configuration. First, the learning rate was not uniform across the network: the BAA module was assigned three times the base rate, since its bias term was randomly initialized, whereas the pretrained encoder was scaled to 0.05 times that rate. Second, both validation and final evaluation were performed on the exponential moving average (EMA) weights rather than on the raw parameters.

\begin{table}[!t]
\centering
\caption{\textbf{Training configuration.} PyTorch, single NVIDIA A100.}
\label{tab:hyperparams}
\renewcommand{\arraystretch}{1.12}
\setlength{\tabcolsep}{3pt}
\footnotesize
\begin{tabular*}{\columnwidth}{@{\extracolsep{\fill}}ll@{}}
\toprule
Setting & Value \\
\midrule
Optimizer & AdamW \\
Peak learning rate & $5 \times 10^{-4}$ \\
Weight decay & $1 \times 10^{-4}$ \\
LR multiplier, BAA & $3.0\times$ base \\
LR multiplier, encoder & $0.05\times$ base \\
Schedule & cosine annealing \\
Warmup & linear, first 10\% of epochs \\
Max epochs & 300 \\
Batch size (train / eval) & 64 / 16 \\
Early stopping patience & 15 epochs \\
EMA decay & 0.999 (EMA weights used for evaluation) \\
Input resolution & $384 \times 384$ \\
\midrule
Dice : BCE & 1 : 1 \\
Deep supervision weights & $[1, 0.5, 0.25, 0.125, 0.0625]$ \\
Label smoothing $\epsilon_{ls}$ & 0.1 \\
MSAR stage weights $\{w_l\}$ & $[0.6, 0.22, 0.13, 0.05]$ \\
$\omega_{seg} : \omega_{cls} : \omega_{align}$ & 1 : 3 : 1 \\
Decision threshold & 0.5 (both tasks) \\
\bottomrule
\end{tabular*}
\end{table}

\subsubsection{Evaluation Metrics}Segmentation quality was measured with five metrics: the Dice Similarity Coefficient (DSC), the Intersection over Union (IoU), sensitivity (SE), specificity (SP), and accuracy (ACC). For classification, each of the two binary tasks (melanoma-vs-rest and seborrheic-keratosis-vs-rest) is evaluated using SE, ACC, SP, F1-score, and AUC, with the mean across the two tasks reported as the overall classification performance.

The validation cohort drove both model selection and checkpointing. The retained checkpoint was the one that maximized a composite criterion, the mean of the segmentation Dice coefficient and the classification AUC.

\begin{table*}[t]
\centering
\caption{Ablation study of different components on the ISIC 2017 dataset.}
\label{tab:ablation}
\renewcommand{\arraystretch}{1.1}
\setlength{\tabcolsep}{3pt}
\small
\resizebox{\textwidth}{!}{
\begin{tabular}{lcccccc|cccc|cccc}
\toprule
\multirow{2}{*}{Model} 
& \multirow{2}{*}{BAA} 
& \multirow{2}{*}{iECA} 
& \multirow{2}{*}{MSSA} 
& \multirow{2}{*}{MSAR} 
& \multirow{2}{*}{Params} 
& \multicolumn{4}{c|}{Segmentation Metrics} 
& \multicolumn{4}{c}{Classification Metrics} \\
\cmidrule(lr){7-10} \cmidrule(lr){11-14}
& & & & & (M) 
& DSC & IoU & SE & SP 
& AUC & F1 & SE & SP \\
\midrule

I: Baseline
& — & — & — & — & 33.97
& 83.76$\pm$0.94 & 75.09$\pm$1.21 & 81.02$\pm$2.69 & 99.02$\pm$0.29
& 85.34$\pm$1.58 & 57.14$\pm$5.81 & \textbf{64.96}$\pm$9.25 & 85.51$\pm$8.07 \\

II: +BAA
& $\checkmark$ & — & — & — & 33.97
& 84.84$\pm$0.58 & 76.51$\pm$0.84 & 81.47$\pm$1.54 & 99.18$\pm$0.14
& 84.48$\pm$1.78 & 59.41$\pm$0.97 & 56.13$\pm$3.56 & 92.06$\pm$1.69 \\

III: +iECA 
& $\checkmark$ & $\checkmark$ & — & — & 33.97 
& 84.87$\pm$0.50 & 76.50$\pm$0.51 & 81.00$\pm$0.65 & \textbf{99.21}$\pm$0.06 
& 85.00$\pm$1.65 & \textbf{60.05}$\pm$1.87 & 52.42$\pm$1.31 & \textbf{94.62}$\pm$0.90 \\

IV: +MSSA 
& $\checkmark$ & $\checkmark$ & $\checkmark$ & — & 34.06 
& 84.94$\pm$0.54 & 76.42$\pm$0.78 & 82.38$\pm$1.54 & 98.95$\pm$0.10
& 84.90$\pm$1.44 & 59.18$\pm$2.11 & 61.82$\pm$8.21 & 88.61$\pm$4.31 \\

V: Full model 
& $\checkmark$ & $\checkmark$ & $\checkmark$ & $\checkmark$ & 34.08
& \textbf{85.43}$\pm$0.64 & \textbf{77.16}$\pm$0.79 & \textbf{82.57}$\pm$1.92 & 98.99$\pm$0.37
& \textbf{85.37}$\pm$0.92 & 58.42$\pm$3.83 & 50.71$\pm$5.22 & 94.48$\pm$1.81 \\

\bottomrule
\end{tabular}
}
\end{table*}

\begin{table}[!t]
\centering
\caption{Ablation study of different BAA layer configurations on ISIC2017 dataset.}
\label{tab:baa_ablation}
\renewcommand{\arraystretch}{1.2}
\setlength{\tabcolsep}{3pt}
\footnotesize
\resizebox{\columnwidth}{!}{%
\begin{tabular}{c|l|cc|cc}
\toprule
\multirow{2}{*}{Model} & \multirow{2}{*}{Configuration} 
& \multicolumn{2}{c|}{Segmentation} & \multicolumn{2}{c}{Classification} \\
\cmidrule(lr){3-4} \cmidrule(lr){5-6}
 & & DSC & IoU & AUC & F1 \\
\midrule

I 
& Standard MHA (all layers) 
& 83.76$\pm$0.94 & 75.09$\pm$1.21
& \textbf{85.34}$\pm$1.58 & 57.14$\pm$1.58 \\\

VI 
& BAA (last two layers)
& \textbf{85.22}$\pm$0.53
& \textbf{77.00}$\pm$0.66
& 83.86$\pm$0.82
& 55.10$\pm$2.20 \\

VII 
& BAA (all layers)
& 84.58$\pm$0.79
& 76.24$\pm$1.03 
& 84.40$\pm$1.64
& 55.61$\pm$8.01 \\

II 
& BAA (last layer only) 
& 84.84$\pm$0.58 & 76.51$\pm$0.84 
& 84.48$\pm$1.78 & \textbf{59.41}$\pm$0.97 \\
\bottomrule
\end{tabular}%
}
\end{table}

\begin{table}[!htbp]
\captionsetup{justification=raggedright,singlelinecheck=false}
\caption{Ablation study of spatial alignment strategies on ISIC2017 and PH2 dataset.}
\label{tab:align_ablation}
\centering
\renewcommand{\arraystretch}{1.1}
\setlength{\tabcolsep}{3pt}
\footnotesize
\resizebox{\columnwidth}{!}{%
\begin{tabular}{c|l|cccc}
\toprule
Model & Align Mode & DSC & IoU & AUC & F1 \\
\midrule
\multicolumn{6}{l}{\textit{ISIC2017 (in-domain)}} \\
\midrule
VIII  & None        
& 84.94$\pm$0.54 
& 76.42$\pm$0.78 
& 84.90$\pm$1.44 
& 59.18$\pm$2.11 \\

IX & Deepest     
& 85.15$\pm$0.21 
& 76.87$\pm$0.31 
& \textbf{86.43}$\pm$0.93 
& \textbf{61.39}$\pm$5.08 \\

V   & Multi-scale 
& \textbf{85.43}$\pm$0.64 
& \textbf{77.16}$\pm$0.79 
& 85.37$\pm$1.92 
& 58.42$\pm$3.83 \\

\midrule
\multicolumn{6}{l}{\textit{PH2 (zero-shot)}} \\
\midrule
VIII  & None        
& 90.58$\pm$1.06 
& 83.44$\pm$1.80 
& 91.31$\pm$3.39 
& 65.87$\pm$9.32 \\

IX & Deepest     
& 90.71$\pm$1.41 
& 83.66$\pm$2.38 
& 93.43$\pm$1.81 
& \textbf{75.52}$\pm$9.72 \\

V   & Multi-scale 
& \textbf{91.13}$\pm$0.74 
& \textbf{84.42}$\pm$1.24 
& \textbf{95.14}$\pm$0.48 
& 73.58$\pm$1.39 \\

\bottomrule
\end{tabular}%
}
\end{table}

\begin{table}[!htbp]
\captionsetup{justification=raggedright,singlelinecheck=false}
\caption{Ablation study of different loss weighting strategies on ISIC2017 and PH2 datasets.}
\label{tab:loss_ablation}
\centering
\renewcommand{\arraystretch}{1.1}
\setlength{\tabcolsep}{3pt}
\footnotesize
\resizebox{\columnwidth}{!}{%
\begin{tabular}{c|l|ccc|cc|cc}
\toprule
\multirow{2}{*}{Model} & \multirow{2}{*}{Loss Strategy} 
& \multicolumn{3}{c|}{Weights} 
& \multicolumn{2}{c|}{Segmentation} 
& \multicolumn{2}{c}{Classification} \\
\cmidrule(lr){3-5} \cmidrule(lr){6-7} \cmidrule(lr){8-9}
 & & $w_{\text{seg}}$ & $w_{\text{cls}}$ & $w_{\text{align}}$ 
 & DSC & IoU & AUC & F1 \\
\midrule
\multicolumn{9}{l}{\textit{ISIC2017 (in-domain)}} \\
\midrule
X
& Fixed (1:1) 
& 1.0 & 1.0 & 1.0 
& 85.40$\pm$0.48
& 77.08$\pm$0.60
& 83.76$\pm$1.28
& 56.07$\pm$2.72 \\
XI
& Fixed (1:5) 
& 1.0 & 5.0 & 1.0 
& \textbf{85.48}$\pm$1.05
& \textbf{77.28}$\pm$1.32
& \textbf{85.38}$\pm$0.87
& 56.43$\pm$0.12 \\
XII
& Fixed (1:10) 
& 1.0 & 10.0 & 1.0 
& 85.17$\pm$0.98
& 76.71$\pm$1.25
& 84.35$\pm$1.79
& 54.81$\pm$2.44 \\
V
& Fixed (1:3) 
& 1.0 & 3.0 & 1.0 
& 85.43$\pm$0.64 
& 77.16$\pm$0.79 
& 85.37$\pm$1.92 
& \textbf{58.42}$\pm$3.83 \\

\midrule
\multicolumn{9}{l}{\textit{PH2 (zero-shot)}} \\
\midrule
XI
& Fixed (1:5) 
& 1.0 & 5.0 & 1.0 
& 89.91$\pm$1.67
& 82.41$\pm$2.68
& 93.46$\pm$0.41
& \textbf{77.37}$\pm$0.75 \\
V
& Fixed (1:3) 
& 1.0 & 3.0 & 1.0 
& \textbf{91.13}$\pm$0.74 
& \textbf{84.42}$\pm$1.24 
& \textbf{95.14}$\pm$0.48 
& 73.58$\pm$1.39 \\
\bottomrule
\end{tabular}%
}
\end{table}

\section{Results}
\subsection{Comparison with Other Methods}
\label{subsection:SOTA}
This section benchmarks the proposed model on the ISIC 2017 test set and the PH2 dataset. ISIC 2018 is reported separately on the segmentation task as an additional reference. All reported numbers are averaged over three independent training runs with seeds 42, 1024, and 2024.

\subsubsection{PH2 and ISIC 2018}
To assess cross-dataset generalization, we evaluate MDSkin-Net directly on PH2 (small cohort, $n=200$) and the ISIC 2018 Task~1 test set (larger cohort, $n=1000$) under zero-shot transfer, without any fine-tuning. As shown in Table~\ref{tab:sota_ph2_zeroshot}, on PH2 MDSkin-Net attains a segmentation DSC of 92.28\% with the highest segmentation sensitivity (98.07\%) among all listed methods, and a melanoma-classification AUC of 95.38\%. On ISIC 2018 (Table~\ref{tab:isic2018_zeroshot}), the model reaches a DSC of $88.81\% \pm 0.45$ and an IoU of $81.30\% \pm 0.69$, with all cross-seed standard deviations strictly below $1\%$. The two settings together indicate that MDSkin-Net consistently retains robust performance under zero-shot evaluation across cohorts of varying scales.

\subsubsection{ISIC2017}
We compare MDSkin-Net against representative segmentation and classification methods published over the past decade. Tables~\ref{tab:sota_isic2017_seg} reports the quantitative results. For segmentation, MDSkin-Net achieves a Dice coefficient of 84.63\% and an IoU of 76.16\%, while attaining the highest specificity (99.36\%) among all listed methods. For classification, MDSkin-Net attains a mean AUC of 91.1\% across the two binary tasks of the ISIC 2017 protocol (Mel vs.\ rest and SK vs.\ rest). All results are obtained on the official ISIC 2017 test set without any external training data, while some of the listed baselines follow alternative evaluation protocols, including 7:3 random splits or training with additional ISIC archive images.

\subsection{Ablation Study}
We ablate MDSkin-Net along three axes: the hybrid CNN--Transformer shared encoder, the PAGAM module, and the MSAR. All ablation experiments are run with three random seeds (42, 1024, 2024), and we report mean~$\pm$~standard deviation to reflect the stability of each configuration.

To minimize sources of variance unrelated to the components under study, EMA and ColorJitter were disabled throughout the ablation. Both are model-agnostic enhancements that affect all configurations uniformly, so leaving them enabled would obscure the per-component effects we aim to isolate. As a result, the full model in the ablation setting attains a slightly higher DSC and a slightly lower AUC than the configuration reported in Table~\ref{tab:sota_combined}. This directional split is consistent rather than contradictory. ColorJitter acts as a domain-randomization augmentation: it discourages reliance on ISIC 2017-specific color statistics, which slightly lowers in-domain segmentation fit but improves transfer to cohorts with different color characteristics, raising zero-shot DSC on PH2 (91.13\% to 92.28\%). EMA smooths the weight trajectory and stabilizes the classification decision boundary, so AUC rises in both settings. The only metric that reverses is in-domain DSC, and it does so for the same reason the priors improve cross-cohort transfer: both trade a small amount of in-domain fit for out-of-domain robustness.

\subsubsection{Architecture-Level Ablation}
\label{subsection:Architecture-Level}
Table~\ref{tab:ablation} reports a progressive ablation over five variants on the ISIC 2017 dataset. Model~I serves as the baseline. Models~II--IV incrementally introduce the BAA, iECA, and MSSA modules, respectively, and Model~V completes MDSkin-Net by further imposing the alignment regularization $\omega_{\text{align}}$.

Stacking the structural modules alone (Models~II--IV) reveals a clear task-level trade-off. DSC rises monotonically from 83.76\% to 84.94\%, whereas AUC falls below the baseline (85.34\%) and remains within 84.48\%--85.00\%, which suggests a latent misalignment between the dense-prediction and classification pathways. Incorporating MSAR ($\omega_{\text{align}}$) in Model~V reconciles this conflict: segmentation peaks at DSC 85.43\% and IoU 77.16\%, while classification recovers and surpasses the baseline at an AUC of 85.37\% with a markedly smaller standard deviation. 

\subsubsection{Module-Level Ablation}

To complement the architecture-level study, we further investigate the placement of BAA within the MobileViT bottleneck. Starting from the standard multi-head attention configuration in Model~I, we replace the attention block with BAA in the last layer (Model~II), the last two layers (Model~VI), and all transformer layers (Model~VII) of the bottleneck. Quantitative results are summarized in Table~\ref{tab:baa_ablation}.

\begin{figure*}[t]
    \centering
    \includegraphics[width=\textwidth]{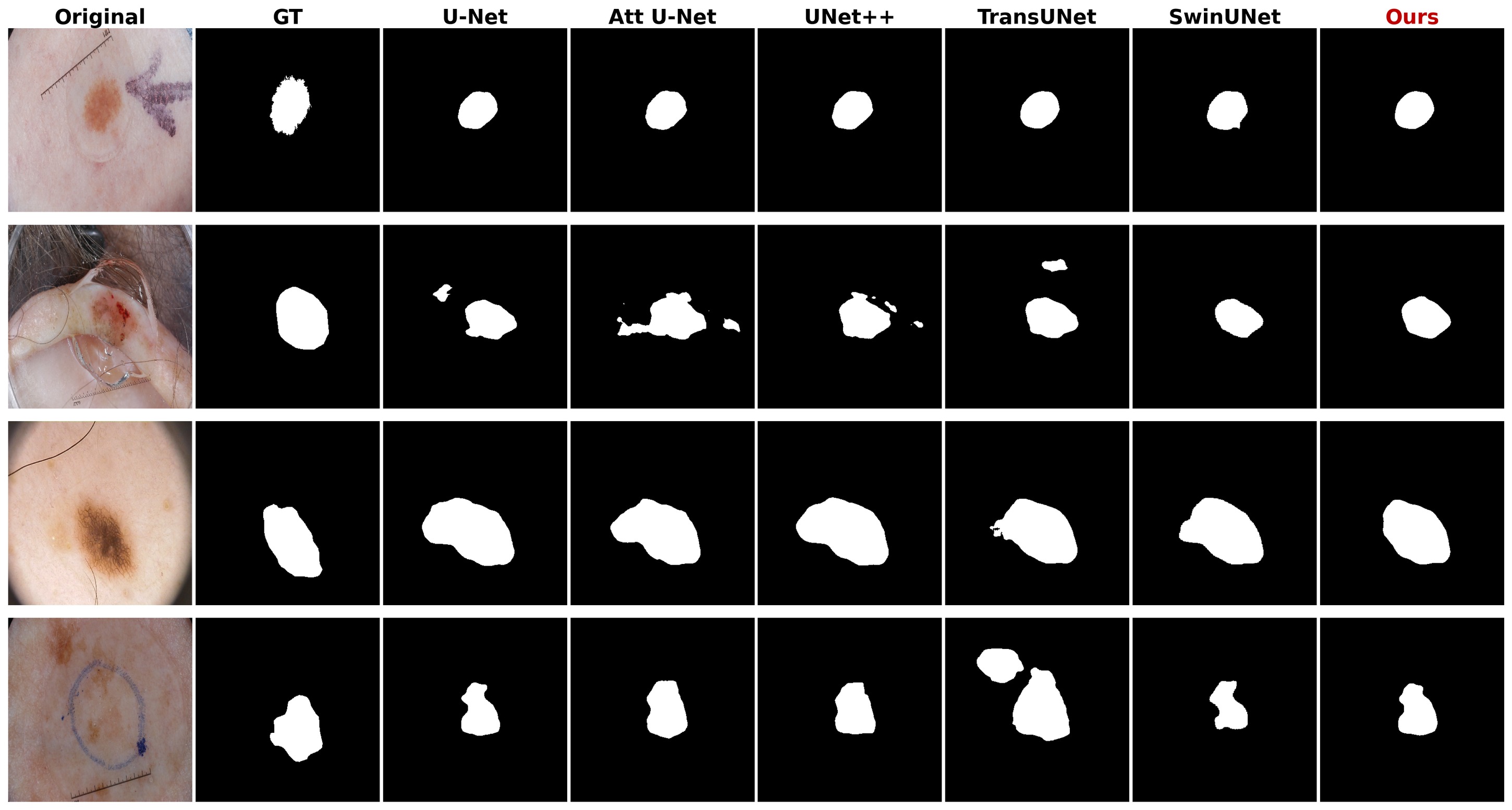}
    \captionsetup{justification=centering}
    \caption{Visualization of segmentation performance between our proposed model and other SOTA methods. From left to right, the columns sequentially display: (1) the original input image; (2) the ground truth mask; (3) the prediction by U-Net; (4) the prediction by Attention U-Net; (5) the prediction by Unet++; (6) the prediction by TransU-Net; (7) the prediction by SwinU-Net; and (8) the segmentation output of our full model.}
    \label{fig:compare}
\end{figure*}

Among the three BAA placements, restricting BAA to the final layer (Model~II) provides the most balanced trade-off. Segmentation performance peaks when BAA replaces the last two layers (Model~VI), achieving a DSC of 85.22\% and an IoU of 77.00\%, whereas extending BAA to all transformer layers (Model~VII) results in a slight regression. Classification exhibits the opposite trend as BAA is applied more broadly: Model~VI records the lowest AUC (83.86\%), while Model~VII shows a markedly larger F1 variance ($\pm 8.01$). In contrast, Model~II achieves the highest AUC (84.48\%) among all BAA variants, while maintaining competitive segmentation performance (DSC 84.84\%, IoU 76.51\%).

\subsubsection{Objective-Level Ablation}

We further ablate the alignment strategy and loss weighting on the ISIC2017 test set and the zero-shot PH2 dataset.

Table~\ref{tab:align_ablation} compares three alignment variants: no alignment (Model~VIII), alignment applied only at the deepest bottleneck (Model~IX), and multi-scale alignment across all encoder stages (Model~V). On the in-domain ISIC2017 test set, the best segmentation and classification performances are achieved by different variants. Model~V leads in segmentation with a DSC of 85.43\% and an IoU of 77.16\%, whereas Model~IX achieves the highest classification performance with an AUC of 86.43\%.

To further examine generalization capability, we evaluate all three variants on the unseen PH2 dataset. Under this cross-dataset setting, Model~V ranks first across DSC (91.13\%), IoU (84.42\%), and AUC (95.14\%), while also exhibiting the lowest standard deviation for all metrics. Table~\ref{tab:loss_ablation} investigates four $\omega_{\mathrm{seg}}{:}\omega_{\mathrm{cls}}$ weighting ratios while fixing $\omega_{\mathrm{align}} = 1.0$. On ISIC2017, performance varies non-monotonically with the weighting ratio and follows a U-shaped trend, peaking at Model~XI ($1{:}5$), which achieves the highest DSC (85.48\%), IoU (77.28\%), and AUC (85.38\%).

Following the same evaluation protocol as the alignment study, we further compare the two best-performing configurations, Model~V ($1{:}3$) and Model~XI ($1{:}5$), on the PH2 dataset. In this cross-dataset evaluation, Model~V consistently outperforms Model~XI on DSC (91.13\% vs.\ 89.91\%), IoU (84.42\% vs.\ 82.41\%), and AUC (95.14\% vs.\ 93.46\%). Therefore, the combination of multi-scale alignment and the $1{:}3$ weighting ratio is adopted in the final MDSkin-Net framework.

\subsection{Visualization}
\subsubsection{Qualitative Examples Alongside Representative Methods}
Fig.~\ref{fig:compare} presents qualitative segmentation results from MDSkin-Net alongside five representative methods (U-Net~\cite{ronneberger2015unet}, Attention U-Net~\cite{oktay2018attention}, U-Net++~\cite{zhou2018unetpp}, TransU-Net~\cite{chen2021transunet}, and SwinU-Net~\cite{cao2022swinunet}). The top two rows are drawn from ISIC 2017, the third from ISIC 2018, and the fourth from PH2. Among the baselines, SwinU-Net produces the masks most comparable to MDSkin-Net, while MDSkin-Net yields visibly smoother lesion boundaries with fewer high-frequency artifacts.

\subsubsection{Visual Ablation Analysis}
\begin{figure*}[t]
    \centering
    \includegraphics[width=\textwidth]{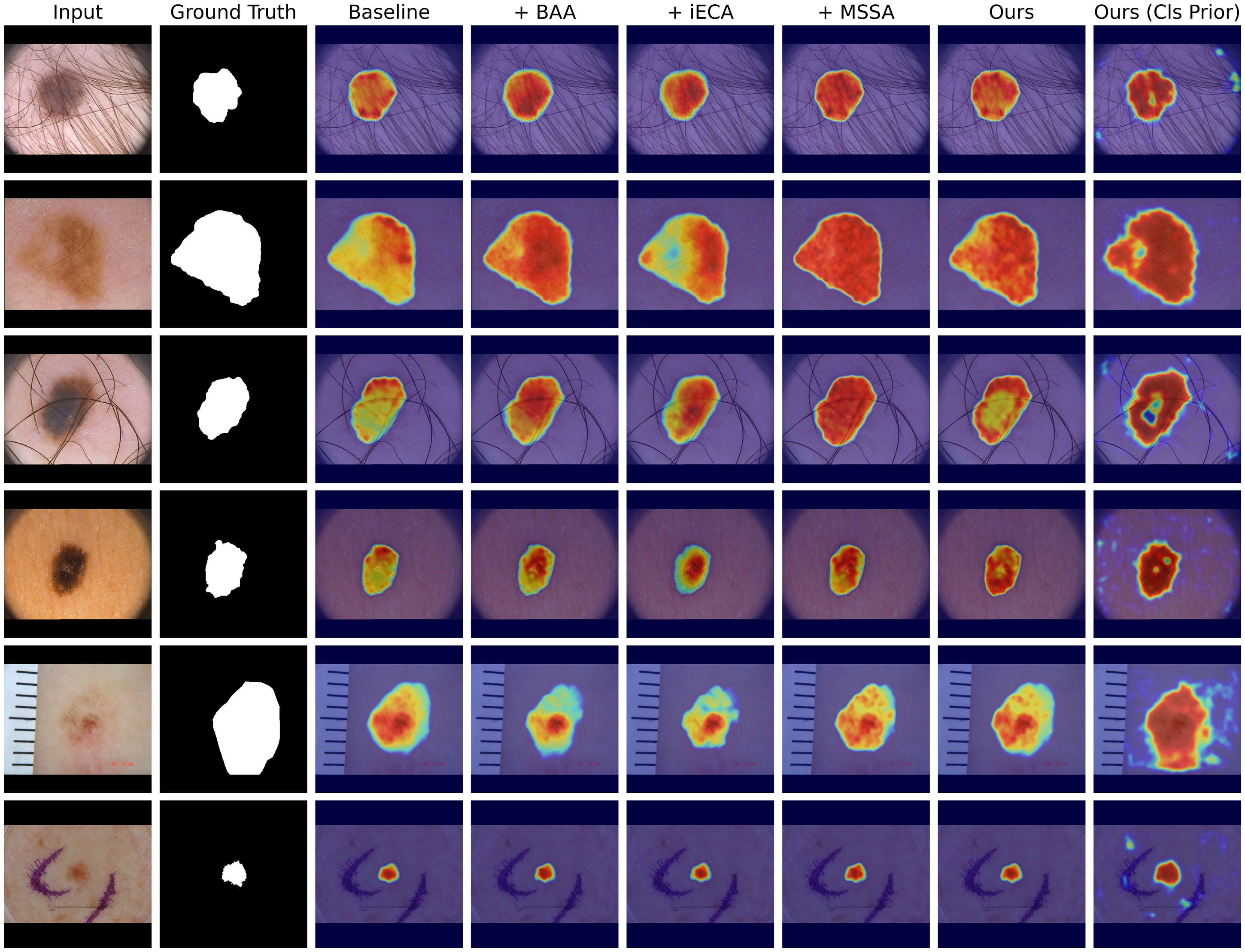}
    \captionsetup{justification=centering}
    \caption{Visual ablation analysis of MDSkin-Net using seed 42. From left to right, the columns sequentially display: (1) the original input image; (2) the ground truth mask; (3) the prediction of the baseline hybrid model; (4) the baseline integrated with the BAA module; (5) the further addition of the iECA module; (6) the inclusion of the MSSA module; (7) the segmentation output of the full architecture; and (8) the learned MSAR spatial prior applied along the classification path}
    \label{fig:ablation_vis}
\end{figure*}

To complement the quantitative results in Table~\ref{tab:ablation}, Fig.~\ref{fig:ablation_vis} provides a visual ablation analysis on samples from three datasets. The top four rows are from PH2, each selected to expose a different dermoscopic condition: a lesion heavily occluded by hair with no annotated dermatologist-recognized cue, a lesion exhibiting the highest level of asymmetry, a lesion presenting a blue-whitish veil, and a lesion exhibiting streaks and pigment structures across multiple spatial scales. The bottom two rows are drawn from the ISIC 2018 Task~1 test set and the ISIC 2017 test set, respectively, both containing extraneous imaging artifacts such as rulers and skin markers.

\section{Discussion}
We approach joint dermoscopic segmentation and classification as a clinical pattern analysis task. Cue-level architectural priors are injected into a CNN-Transformer hybrid backbone to capture abrupt local chromatic transitions, scale-diverse pigment structures, and internal structural asymmetry. The MSAR then couples the two branches, encouraging the classification pathway to attend to lesion-relevant regions under the supervision of the segmentation pathway. Trained solely on the ISIC 2017 training split, MDSkin-Net matches or exceeds its in-domain performance under zero-shot evaluation on PH2 and ISIC 2018.

\subsection{Pattern-Level Priors and Generalization}
On the ISIC 2017 official test set, MDSkin-Net attains a segmentation DSC of 84.72\%, below recent single-task specialists such as SwinUNet (91.83\%) and SVMB-Net (90.56\%). This gap is expected: MDSkin-Net shares one encoder between segmentation and classification and, via MSAR, further ties the classification head to lesion-localized evidence, so part of the encoder capacity serves classification rather than dense prediction alone. Such single-task-accuracy costs under a shared representation are well documented in multi-task learning \cite{vandenhende2022multitask}. It is nonetheless incurred at far lower cost than the strongest single-task baseline (34.06\,M / 18.59\,G vs.\ 45.57\,M / 163.70\,G for SVMB-Net). A counter-intuitive trend emerges from Tables~\ref{tab:sota_combined} and~\ref{tab:isic2018_zeroshot}: under zero-shot evaluation, MDSkin-Net reaches a DSC of 92.38\% on PH2 and 88.92\% on ISIC 2018, both exceeding the in-domain DSC of 84.72\% on ISIC 2017. We attribute this reversal, at least in part, to the inductive bias carried by the three pattern-level priors. iECA normalizes channel-wise statistics, reducing sensitivity to absolute brightness; MSSA operates over multi-dilation receptive fields, reducing sensitivity to structural scale; and BAA anchors attention on saliency-weighted centroids rather than geometric centers, reducing dependence on lesion position. These invariances align with the clinical observation that dermoscopic pigment patterns remain relatively stable across acquisition conditions.

Beyond the pattern-level priors, three design choices are explicitly tailored to the size of the ISIC 2017 training set: the MobileViT bottleneck is restricted to depth 3 to limit the most overfitting-prone Transformer component; MSAR doubles as an implicit regularizer by confining the classification head to lesion-localized evidence, suppressing spurious correlations with background skin or imaging artifacts that may not transfer across cohorts; and label smoothing in the classification head discourages overconfident predictions on the small training distribution. Together with the pattern-level invariances, these capacity-control choices contribute to the observed cross-cohort stability.

Alternative explanations cannot be ruled out: PH2 and ISIC 2018 may also contain visually cleaner images or class distributions favorable to the trained model. We cannot isolate any single component. The above is best read as a hypothesis consistent with current evidence rather than a definitive mechanism; at minimum, the cross-dataset stability indicates that MDSkin-Net has not over-fitted to dataset-specific spurious cues.

\subsection{Spatial Alignment and Task Conflict}
Table~\ref{tab:ablation} reveals a recurring pattern across the first four variants. As BAA, iECA, and MSSA are sequentially added (Models~II--IV), segmentation improves monotonically while classification stays below the baseline (AUC $84.48\%$--$85.00\%$ vs.\ $85.34\%$). This reflects a latent task conflict often encountered in multi-task learning: segmentation rewards local detail, whereas classification benefits from discriminative global semantics.

\begin{figure}[t]
    \centering
    \includegraphics[width=0.8\columnwidth]{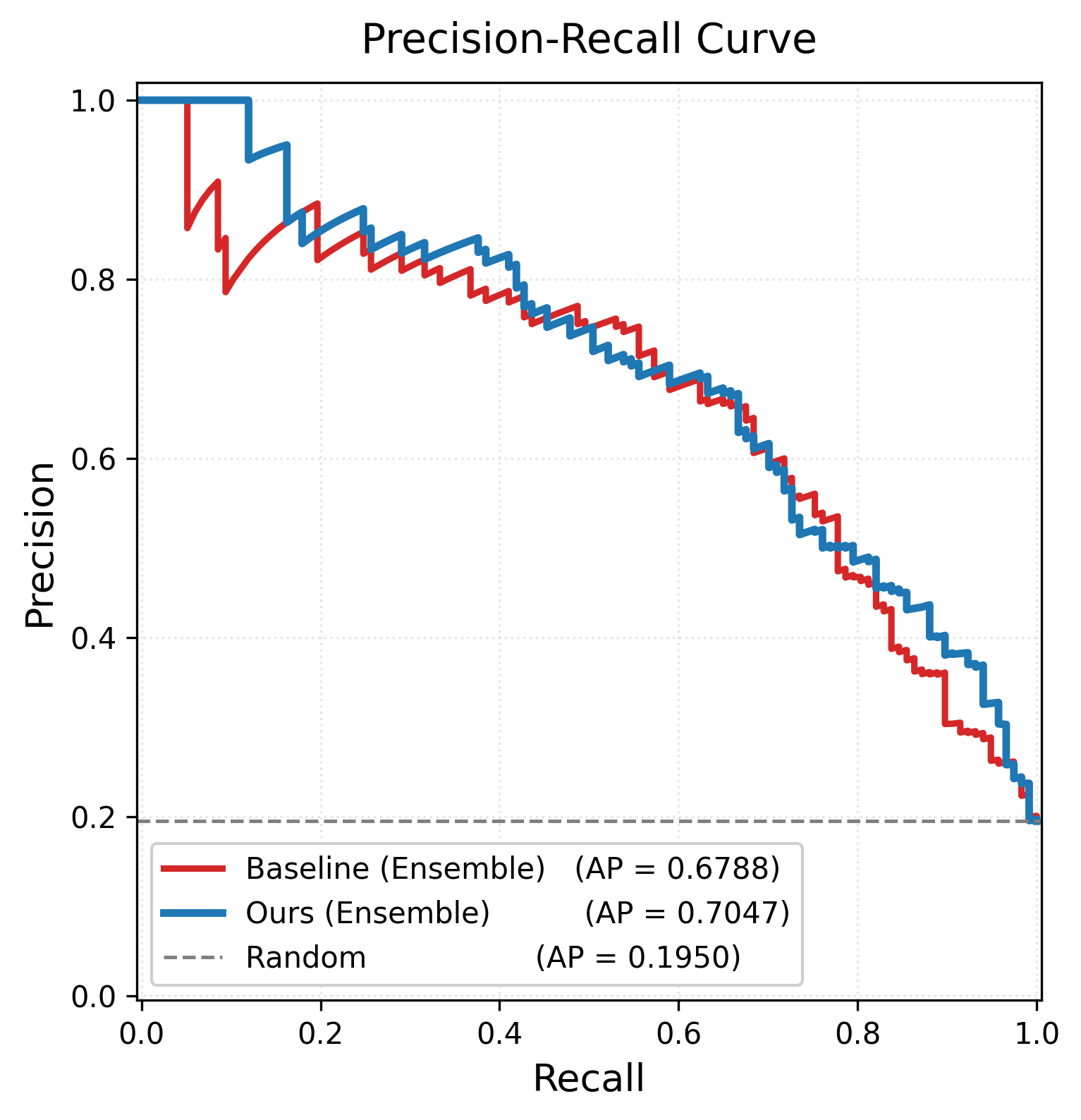}
    \captionsetup{justification=centering}
    \caption{Precision-recall curves comparing the baseline and proposed ensemble models on the ISIC 2017 test set. The proposed architecture achieves a higher AP (0.7047) than the baseline ensemble (0.6788).}
    \label{fig:pr_curve}
\end{figure}

Adding the MSAR in Model~V lifts both metrics to their joint optimum (DSC $85.43\%$, AUC $85.37\%$). Rather than injecting the predicted mask as a hard prior, the alignment uses a residual gate $1 + \alpha \sigma(A)$ with a learnable $\alpha$ (initialized at $0.1$), where $A$ denotes the alignment map. This formulation preserves the original classification pathway and lets the network learn how strongly segmentation-derived priors should modulate classification features, providing a soft coupling rather than a hard injection.

Fig.~\ref{fig:ablation_vis} (rightmost column) further supports this mechanism. The learned MSAR spatial prior is applied along the classification path before the classification head, and concentrates on the lesion region across hierarchical scales. This indicates that classification decisions are encouraged to draw from multi-scale lesion structure rather than from background context.

\subsection{Layer-Specific Prior Placement}
Table~\ref{tab:baa_ablation} indicates that BAA performs best when restricted to the final transformer layer (Model~II). Extending it to the last two layers (Model~VI) maximizes segmentation at the cost of classification (AUC $83.86\%$), and applying it to all transformer layers (Model~VII) reverses the segmentation trend and inflates the F1 variance to $\pm 8.01$. This pattern is consistent with the structural role of BAA as a high-level prior over saliency-weighted patch geometry. Although every transformer layer captures global dependencies through MSA, shallow layers operate on features that have not yet been semantically distilled; applying patch-wise asymmetry at this stage imposes the bias on noisier descriptors and can propagate spurious patch correlations. At the deepest bottleneck stage, the feature map is heavily downsampled and semantically condensed, so the same computation operates on cleaner, higher-level features and yields a more reliable asymmetry signal without resorting to hard mask injection. The same logic applies clinically, since Pattern Analysis itself is a high-level diagnostic cue: asymmetry is assessed only after the underlying dermoscopic structures have been recognized. We therefore place BAA at the bottleneck, while iECA and MSSA, which modulate mid-level features, are inserted along the skip connections. Taken together, these placements suggest a simple design principle: the depth at which a clinical prior is injected should match its semantic level.

\subsection{Operating Point and Clinical Use}
On the ISIC 2017 test set, the proposed ensemble achieves an average precision of 70.47\%, outperforming the baseline ensemble (67.88\%) against a random-classifier baseline of 19.50\% that reflects the positive class prevalence (Fig.~\ref{fig:pr_curve}). At the default decision threshold of 0.5, this corresponds to a sensitivity of 52.1\% and a specificity of 94.8\%, a conservative operating point that trades recall for precision and prioritizes the suppression of false positives over the capture of every malignant case.

The precision-recall curve indicates that this trade-off is 
adjustable rather than fixed. At a recall of approximately 0.7, 
precision remains around 0.55, still nearly threefold above the 
random baseline. The operating point can therefore be shifted 
toward higher sensitivity when clinically warranted, without 
collapsing into noise. We therefore position MDSkin-Net as a triage tool for high-throughput pre-screening followed by expert review, rather than a substitute for dermatopathological diagnosis. Under this deployment, a high specificity at the default threshold reduces the dermatologist workload on clearly negative cases, while ambiguous or low-confidence predictions remain available for closer inspection.

\subsection{Limitations and Future Directions}
\subsubsection{Model-Level Limitations}
The evaluated datasets are limited in scale and lack the long-tail distributions of real-world clinical practice. To prevent overfitting, we restrict Transformer components to the bottleneck and keep the encoder predominantly convolutional; larger multi-center datasets would enable safer expansion of the Transformer share for more complex lesion differentiation.

MSAR mitigates the latent task conflict only at the spatial level by constraining classification attention to lesion-localized evidence, without directly addressing gradient-level conflicts in the loss space. Adaptive optimization strategies such as dynamic gradient projection or uncertainty-based loss weighting may offer a more principled remedy.

\subsubsection{Clinical-Level Limitations}
Single-image diagnosis departs from authentic dermatological practice, which often relies on the longitudinal evolution of a lesion. Static datasets evaluate lesions in isolation and omit this temporal dimension, which motivates future work on time-series modeling over patient histories rather than cross-sectional snapshots.

Our framework is trained on dermoscopic images acquired under standardized conditions. Broader deployment scenarios such as total-body photography introduce distortion, scale variation, and lighting-induced color shifts that may degrade the cues used by PAGAM. This requires reproducible imaging protocols in tandem with algorithmic advances.

Finally, the classification labels in ISIC 2017 and PH2 are derived 
from clinical and dermoscopic assessment rather than histopathology. 
Dermoscopic diagnosis itself exhibits substantial inter-observer 
variability, with sensitivity and specificity differing markedly 
between expert and less-experienced evaluators 
\cite{tschandl2019comparison}. These label-level uncertainties 
propagate into any model trained on such cohorts, including ours, so 
the reported AUC should be interpreted as agreement with expert 
consensus rather than with biological ground truth. Prospective 
cohorts with histology-confirmed labels remain necessary for 
definitive clinical validation.

\section{Conclusion}
We presented MDSkin-Net, an end-to-end multi-task framework for skin lesion segmentation and classification that integrates a hybrid CNN-Transformer encoder, a Pattern Analysis-Guided Attention Module, and a multi-scale spatial alignment regularization. Trained solely on the ISIC 2017 training set, the ensemble model transfers robustly under zero-shot evaluation, achieving an AUC of 97.84\% for melanoma classification and a DSC of 92.38\% on PH2, and a DSC of 88.92\% on the ISIC 2018 Task 1 test set. On the in-domain ISIC 2017 benchmark, the ensemble attains a mean AUC of 91.60\% across the two binary classification tasks defined by the challenge protocol and a DSC of 84.72\% for segmentation, remaining competitive with prior baselines despite the absence of external dermoscopy data. Together, these results suggest that cue-level Pattern Analysis priors yield representations that transfer consistently across cohorts of different scales. Future work will address multi-task optimization challenges and, in active collaboration with clinical dermatologists, extend the framework toward longitudinal, histology-confirmed evaluation.

\bibliographystyle{IEEEtran}
\bibliography{refs}

\end{document}